%% file: main.tex
\documentclass[10pt,twocolumn,letterpaper]{article}
\usepackage[accsupp]{axessibility}  

\usepackage[pagenumbers]{cvpr}  

\usepackage{graphicx}
\usepackage{capt-of}
\usepackage{xcolor}
\usepackage{amsmath,amssymb,amsfonts,dsfont,pifont,bm,bbm,mathrsfs,mathtools,nicefrac}
\usepackage{algorithm,algpseudocode,listings}
\usepackage{booktabs,multirow,adjustbox,diagbox,threeparttable}
\definecolor{citeblue}{RGB}{48,111,186}
\usepackage[pagebackref,breaklinks,colorlinks,citecolor=citeblue,bookmarks=false]{hyperref}
\usepackage[capitalize]{cleveref}  
\crefname{section}{Sec.}{Secs.}
\Crefname{section}{Section}{Sections}
\crefname{table}{Tab.}{Tabs.}
\Crefname{table}{Table}{Tables}
\crefname{figure}{Fig.}{Figs.}
\Crefname{figure}{Figure}{Figures}
\crefname{equation}{Eq.}{Eqs.}
\Crefname{equation}{Equation}{Equations}
\newcommand{\tocite}[1]{\textcolor{red}{[TO CITE]}}
\newcommand{\toref}[1]{\textcolor{red}{[TO REF]}}
\newcommand{\method}{\textcolor{black}{PoF3D}\xspace}
\newcommand{\supp}{\textit{Supplementary Material}\xspace}

\newcommand{\D}{\mathrm{D}}

\def\cvprPaperID{0}
\def\confName{CVPR}
\def\confYear{2023}

\newcommand\nonumfootnote[1]{%
\begingroup%
    \renewcommand\thefootnote{}\footnote{\hspace{-3.7pt}#1}%
    \addtocounter{footnote}{-1}%
\endgroup%
}

\begin{document}

\title{Learning 3D-aware Image Synthesis with Unknown Pose Distribution}

\author{
  Zifan Shi\textsuperscript{$\dagger$*1} \quad
  Yujun Shen\textsuperscript{$\dagger$2} \quad
  Yinghao Xu\textsuperscript{*3} \quad
  Sida Peng\textsuperscript{4} \quad
  Yiyi Liao\textsuperscript{4} \quad
  Sheng Guo\textsuperscript{2} \\
  Qifeng Chen\textsuperscript{1} \quad
  Dit-Yan Yeung\textsuperscript{1} \\[5pt]
  \textsuperscript{1}HKUST
  \quad \textsuperscript{2}Ant Group
  \quad \textsuperscript{3}CUHK 
  \quad \textsuperscript{4}Zhejiang University  \\
}

\maketitle

\input{sections/0.abs.tex}
\input{sections/1.intro.tex}

\input{sections/2.related.tex}
\input{sections/3.method.tex}
\input{sections/4.exp.tex}

\input{sections/5.conclusion.tex}
\input{sections/6.ref.tex}
\input{supp/0.overview.tex}
\input{supp/1.training.tex}
\input{supp/2.baselines.tex}

\input{supp/3.results.tex}
\input{supp/4.discussions.tex}

\end{document}

%% file: sections/0.abs.tex
\begin{abstract}

Existing methods for 3D-aware image synthesis largely depend on the 3D pose distribution pre-estimated on the training set.
An inaccurate estimation may mislead the model into learning faulty geometry.
This work proposes \textbf{\method} that frees generative radiance fields from the requirements of 3D pose priors.
We first equip the generator with an efficient pose learner, which is able to infer a pose from a latent code, to approximate the underlying true pose distribution automatically.
We then assign the discriminator a task to learn pose distribution under the supervision of the generator and to differentiate real and synthesized images with the predicted pose as the condition.
The pose-free generator and the pose-aware discriminator are jointly trained in an adversarial manner.
Extensive results on a couple of datasets confirm that the performance of our approach, regarding both image quality and geometry quality, is on par with state of the art.
To our best knowledge, \method demonstrates the feasibility of learning high-quality 3D-aware image synthesis without using 3D pose priors for the first time.
Project page can be found \href{https://vivianszf.github.io/pof3d/}{here}.
%
\nonumfootnote{$\dagger$ indicates equal contribution.\\
\indent {\hspace{1mm}}* This work was done during an internship at Ant Group.}

\end{abstract}
\vspace{-15pt}

%% file: sections/1.intro.tex
\section{Introduction}\label{sec:intro}

3D-aware image generation has recently received growing attention due to its potential applications~\cite{graf,giraffe,pigan,volumegan,stylenerf,eg3d,depthgan}.
Compared with 2D synthesis, 3D-aware image synthesis requires the understanding of the geometry underlying 2D images, which is commonly achieved by incorporating 3D representations, such as neural radiance fields (NeRF)~\cite{nerf, occupancy, deepsdf, neuralbody, nerf++, mipnerf}, into generative models like generative adversarial networks (GANs)~\cite{gan}.
Such a formulation allows explicit camera control over the synthesized results, which fits better with our 3D world.

\begin{figure}[t]
\begin{center}
\includegraphics[width=1\linewidth]{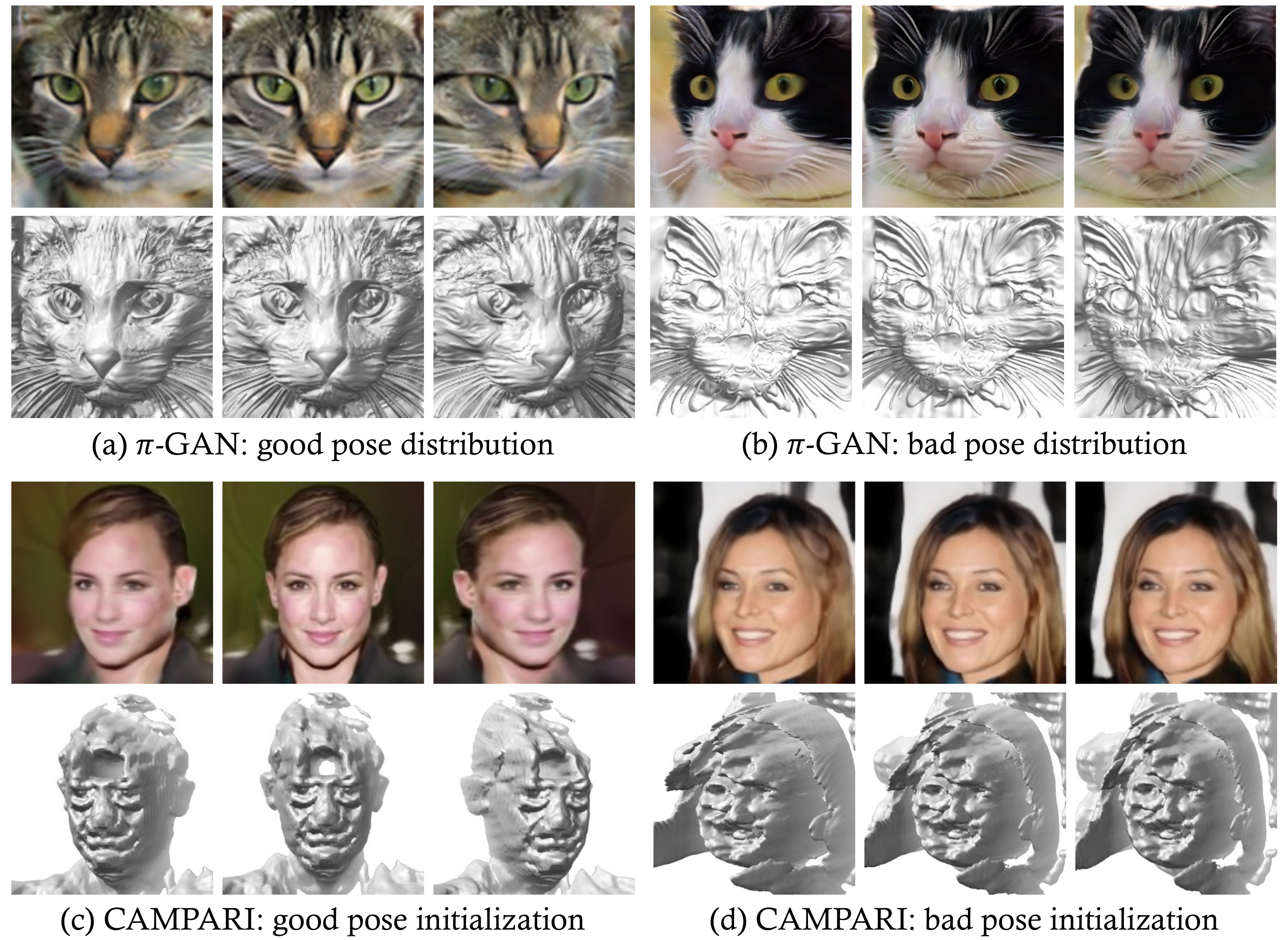}
\end{center}
\vspace{-20pt}
\caption{
    \textbf{Sensitivity to pose priors} in existing 3D-aware image synthesis approaches.
    $\pi$-GAN~\cite{pigan} works well given an adequate pose distribution in (a), but fails to learn decent geometries given a \textit{slightly changed} distribution in (b).
    CAMPARI~\cite{campari} delivers reasonable results relying on a good initial pose distribution in (c), but suffers from a wrongly estimated initialization in (d).
}
\label{fig:teaser}
\vspace{-15pt}
\end{figure}

To enable 3D-aware image synthesis from 2D image collections, existing attempts usually rely on adequate camera pose priors~\cite{graf,giraffe,pigan,volumegan,stylenerf,eg3d,campari} for training.  
The priors are either estimated by conducting multiple pre-experiments~\cite{pigan,volumegan,stylenerf}, or obtained by borrowing external pose predictors or annotations~\cite{eg3d}.
Besides using a fixed distribution for pose sampling, some studies also propose to tune the pose priors together with the learning of image generation~\cite{campari}, but they are still dependent on the initial pose distribution.
Consequently, although previous methods can produce satisfying images and geometry, their performance is highly sensitive to the given pose prior.
For example, on the Cats dataset~\cite{cats}, $\pi$-GAN~\cite{pigan} works well with a uniform pose distribution [-0.5, 0.5] (\cref{fig:teaser}a), but fails to generate a decent underlying geometry when changing the distribution to [-0.3, 0.3] (\cref{fig:teaser}b).
Similarly, on the CelebA dataset~\cite{celeba}, CAMPARI~\cite{campari} learns 3D rotation when using Gaussian distribution $\mathcal{N}(0, 0.24)$ as the initial prior (\cref{fig:teaser}c), but loses the canonical space (\textit{e.g.}, the middle image of \cref{fig:teaser}d should be under the frontal view) when changing the initialization to $\mathcal{N}(0, 0.12)$ (\cref{fig:teaser}d). 
Such a reliance on pose priors causes unexpected instability of 3D-aware image synthesis, which may burden this task with heavy experimental cost.

In this work, we propose a new paradigm for 3D-aware image synthesis, which removes the requirements for pose priors.
Typically, a latent code is bound to the 3D content alone, where the camera pose is independently sampled from a manually designed distribution.
Our method, however, maps a latent code to an image, which is implicitly composed of a 3D representation (\textit{i.e.}, neural radiance field) and a camera pose that can render that image.
In this way, the camera pose is directly inferred from the latent code and jointly learned with the content, simplifying the input requirement.
To facilitate the pose-free generator better capturing the underlying pose distribution, we re-design the discriminator to make it pose-aware.
Concretely, we tailor the discriminator with a pose branch which is required to predict a camera pose from a given image.
Then, the estimated pose is further treated as the conditional pseudo label when performing real/fake discrimination.
The pose branch in the discriminator learns from the synthesized data and its corresponding pose that is encoded in the latent code, and in turn use the discrimination score to update the pose branch in the generator.
With such a loop-back optimization process, two pose branches can align the fake data with the distribution of the dataset.
We evaluate our \textit{pose-free} method, which we call \textbf{\method} for short, on various datasets, including FFHQ~\cite{stylegan}, Cats~\cite{cats}, and Shapenet Cars~\cite{shapenet}. 
Both qualitative and quantitative results demonstrate that \method frees 3D-aware image synthesis from hyper-parameter tuning on pose distributions and dataset labeling, and achieves on par performance with state-of-the-art in terms of image quality and geometry quality.

%% file: sections/2.related.tex
\section{Related Work}\label{sec:related-work}

\noindent{\textbf{3D-aware Image Synthesis}}. 
3D-aware image synthesis has achieved remarkable success recently~\cite{shi2022deep,xia2022survey}.
Different from 2D synthesis~\cite{stylegan,stylegan2,stylegan3,ghfeat}, VON~\cite{von}, HoloGAN~\cite{hologan}, and BlockGAN~\cite{nguyen2020blockgan} propose to adopt voxels as 3D representation for image rendering but suffer from the poor image quality and consistency due to the voxel resolution restriction. 
Then a series of works~\cite{graf, pigan, gof, shadegan, gram} introduce neural implicit function~\cite{nerf, occupancy, deepsdf} as the underlying 3D representation for image rendering.
However, rendering high-resolution images with direct volume rendering is very heavy.
Lots of works~\cite{giraffe, stylenerf, eg3d, volumegan, stylesdf, voxgraf, epigraf, stylempi, xu2022discoscene} resort to either 2D convolutional upsamplers~\cite{giraffe, stylenerf, eg3d, volumegan, stylesdf}, multi-plane images rendering~\cite{stylempi}, sparse-voxel~\cite{voxgraf} inference, patch-based training~\cite{epigraf} to speed up training and inference.
Besides, the geometry quality is not promised and some works~\cite{shadegan,shi2022improving} focus on the improvement of geometry quality.
Although these methods are able to synthesize high-quality and 3D-consistent images, they are restricted to strong pose prior, ~\textit{i.e.}, manually tuned pose distribution, or well-annotated camera poses~\cite{eg3d}.
Our work instead naturally enables 3D-aware image synthesis without any pose prior via  a pose-free generator, which gives a fruitful avenue for future work. 

\noindent{\textbf{Camera Learning from 2D Images}}.
Learning neural radiance fields (NeRF)~\cite{nerf} requires accurate pose annotations.
To overcome this, some works~\cite{wang2021nerfmm, kuang2022neroic} try to optimize the annotated camera parameters or directly estimate camera poses with a small network, which is conceptually comparable to ours.
On the other hand, our work differs from theirs in that our method focuses on learning pose distribution of a real dataset rather than exact camera poses.
CAMPARI~\cite{campari} and 3DGP~\cite{3DGP} investigate a similar problem, aiming to estimate the pose distribution of an actual dataset.
However, both of them need a manually designed prior,~\textit{i.e.}, uniform or Gaussian, and fail to learn appropriate 3D geometry when the prior is very distant from the native distribution.
In contrast, our work abandons all the manually designed prior and allows the generator to learn pose distribution automatically via adversarial training.

\definecolor{myorange}{RGB}{222,131,68}
\begin{figure*}[t]
\begin{center}
\includegraphics[width=1\linewidth]{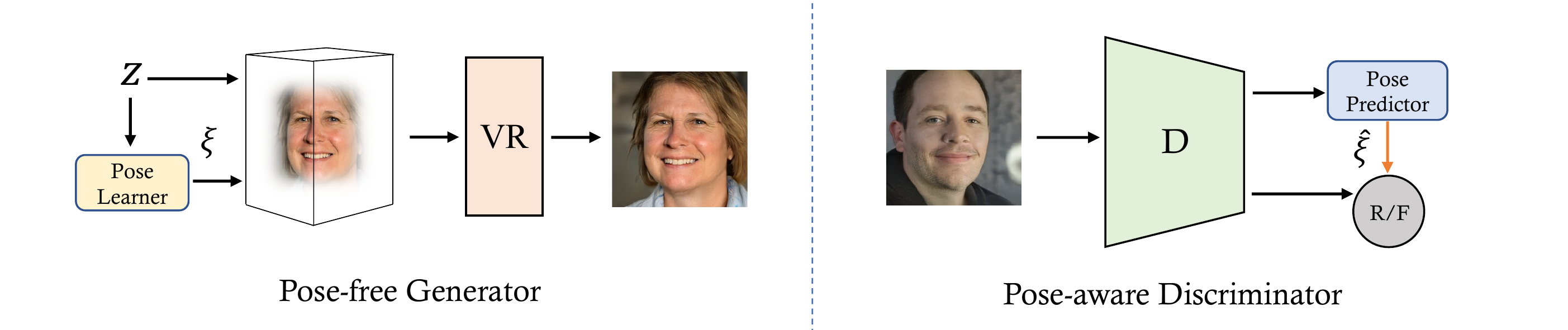}
\end{center}
\vspace{-20pt}
\caption{
    \textbf{Framework of PoF3D}, which consists of a pose-free generator and a pose-aware discriminator. The pose-free generator maps a latent code to a neural radiance field as well as a camera pose, followed by a volume renderer (VR) to output the final image. The pose-aware discriminator first predicts a camera pose from the given image and then use it as the pseudo label for conditional real/fake discrimination, indicated by the \textbf{\textcolor{myorange}{orange}} arrow.
}
\vspace{-10pt}
\label{fig:framework}
\end{figure*}

%% file: sections/3.method.tex
\section{Method}\label{sec:method}
In this work, we propose a new paradigm for 3D-aware image synthesis to free the model from the requirement of pose priors.
To achieve this goal, we re-design both the generator and the discriminator in conventional 3D-aware GANs~\cite{graf, pigan, stylenerf, eg3d}.
Specifically, the pose-free generator is equipped with a pose learner that infers the camera pose from the latent code. 
The pose-aware discriminator extracts a pose from the given image and uses it as the conditional label when performing real/fake classification.
The framework is shown in \cref{fig:framework}.
Before going into details, we first briefly introduce the generative neural radiance field, which plays a crucial role in 3D-aware image synthesis.

\subsection{Preliminary}
A neural radiance field (NeRF)~\cite{nerf}, $\mathrm{F}(\mathbf{x}, \mathbf{d}) \rightarrow (\mathbf{c}, \sigma)$, provides color $\mathbf{c} \in \mathbb{R}^3$ and volume density $\sigma \in \mathbb{R}$ from a coordinate $\mathbf{x} \in \mathbb{R}^3$ and a viewing direction $\mathbf{d} \in \mathbb{S}^2$, typically parameterized with multi-layer perceptron (MLP) networks.
Then, the pixel values are accumulated from the colors and densities of the points on the sampled rays.
However, NeRF is highly dependent on multi-view supervision, leading to the inability to learn from a single-view image collection.
To enable random sampling from single-view captures, recent attempts propose to condition NeRF with a latent code $\mathbf{z}$, resulting in their generative forms~\cite{graf, pigan}, $\mathrm{G}(\mathbf{x}, \mathbf{d}, \mathbf{z}) \rightarrow (\mathbf{c}, \sigma)$, to achieve diverse 3D-aware generation.

\subsection{Pose-free Generator}

Different from NeRF, the camera pose $\xi$ deriving the spatial point $\mathbf{x}$ and view direction $\mathbf{d}$ in conventional 3D-aware generators is randomly sampled from a prior distribution $p_{\xi}$ rather than well annotated by MVS~\cite{MVS} and SfM~\cite{SFM}.
Such a prior distribution requires the knowledge of the pose distribution, and it should be tailored for different datasets, which introduces non-trivial hyperparameter tuning for model training.
This generation process can be formulated as 
\begin{align}
\mathrm{G}(\mathbf{z}, \xi) = \mathbf{I}_f \sim p_{\theta}(\mathbf{I}_f|\mathbf{z}, \xi), \label{eq:cond-v1}
\end{align}
where the generator $\mathrm{G}$ synthesizes the image $\mathbf{I}_f$ by modeling the conditional probability given the latent code and the camera pose.
Since the camera pose is independent of the latent code, we can rewrite \cref{eq:cond-v1} in the following form:
\begin{align}
       p_{\theta}(\mathbf{I}_f|\mathbf{z}, \xi) 
                               = \frac{p_{\theta}(\mathbf{I}_f, \mathbf{z}, \xi)}{p(\mathbf{z}, \xi)} 
                               = \frac{p_{\theta}(\mathbf{I}_f, \xi | \mathbf{z})}{p(\xi)} = \frac{p_{\theta}(\mathbf{I}_f, \xi | \mathbf{z})}{p_{\xi}}.
\end{align}
Obviously, if the pose prior does not align well with the real distribution, it is hard to correctly estimate the image distribution $p_{\theta}(\mathbf{I}_f| \mathbf{z}, \xi)$.
As noticed by~\cite{volumegan, stylenerf}, it always makes the training diverge if the prior does not align well with the real distribution. 
To free the generator from sampling poses from a prior distribution, we follow the formulation of conventional 2D GANs that synthesizes images from the latent code only:
\begin{align}
\mathrm{G}(\mathbf{z}, \Psi(\mathbf{z})) = \mathbf{I}_f \sim p_{\theta}(\mathbf{I}_f |\mathbf{z}, \Psi(\mathbf{z})), \label{eq:nocond}
\end{align}
where the camera pose $\xi$ is parameterized with a nonlinear function $\Psi(\cdot)$ that takes $\mathbf{z}$ as input.
Concretely, we implement $\Psi(\cdot)$ by introducing an additional pose branch on the top of the generator.
The estimated camera pose is further fed into a generative radiance field to render a 2D image.
Compared with previous solutions, our generator aims at approximating the conditional probability only from the latent observation, which is prone to simulating the native data distribution without taking the pose prior into account.

\subsection{Pose-aware Discriminator}\label{sec:method:discriminator}

As mentioned above, we remove the pose prior of a 3D-aware generator by parameterizing camera pose $\xi$ from a latent code $\mathbf{z}$.
The 3D representation leveraged in the generator is helpful in disentangling the camera factor.
However, the conventional discriminator remains to differentiate real and fake images from 2D space, leading to inadequate supervision to factorize camera poses from latent codes.
It easily makes the generator synthesize flat shapes and learn invalid camera distribution.
%
Therefore, we propose to make the discriminator pose-aware.

\noindent{\textbf{Learning Pose-aware Discriminator.}}
To let the discriminator become aware of poses, we assign a pose estimation task on \textit{fake images} beyond bi-class domain classification, which is to derive the pose information from the given images.
We introduce a pose branch $\Phi(\cdot)$ on top of the discriminator and optimize it with the following objective:
\begin{align}
    \hat{\xi} &= \Phi(\mathbf{I}_f), \\ 
    \mathcal{L}_{pose} &= l_2(\hat{\xi}, \xi), \label{eq:pose-loss}
\end{align}
where $l_2(\cdot)$ denotes the function measuring $l_2$ distance between poses. $\xi$ is the pose that is inferred from the same latent code as $\mathbf{I}_f$ by the generator.

\noindent{\textbf{Performing Pose-aware Discrimination}}.
With the help of the pose branch, our discriminator can be adopted to infer camera poses for the given images.
Even though the pose branch training is done on fake images, it is also generalizable on real images $\mathbf{I}_r$ from the dataset.
In such cases, we can leverage the pose-aware discriminator to perform discrimination on real or fake images as well as differentiate whether the image is attached with an accurate pose or not.
Concretely, we first ask the discriminator to extract a pose from the input image, and then the pose is treated as a pseudo label to perform conditional bi-class classification.
Here, we take the real image part of the discriminator loss as an example:
\begin{align}
    \mathcal{L}_\mathrm{D}^{\mathbf{I}_r}  = - \mathbb{E}[\log(\mathrm{D}(\mathbf{I}_r | \ \Phi(\mathbf{I}_r))].
\end{align}
In this way, we facilitate the discriminator with awareness of pose cues, leading to better pose alignment across various synthesized samples.
As a result, this alignment also implicitly prevents the generator from degenerate solutions where a flat shape is usually generated with huge artifacts. 
It is worth noting that the discriminator learns poses from the generator and, in turn, uses the conditional discrimination score to update the pose branch in the generator. Therefore, the poses are learned in a loop-back manner without any annotations.

\subsection{Training Objectives}

\noindent\textbf{Adversarial Loss.}
We use the standard adversarial loss for training following~\cite{gan},
\begin{align}
    \mathcal{L}_\mathrm{D}  =& - \mathbb{E}\big[\log\big(1 - \D(\mathbf{I}_f| \Phi(\mathbf{I}_f))\big)\big] \nonumber \\
    & - \mathbb{E}\big[\log\big(\D(\mathbf{I}_r | \Phi(\mathbf{I}_r))\big)\big] + \lambda \mathbb{E}\big[||\nabla_{\mathbf{I}_r }\D(\mathbf{I}_r)||_2^2\big], \label{eq:d_loss} \\ 
    \mathcal{L}_\mathrm{G}  = &- \mathbb{E}[\log(\D(\mathbf{I}_f| \Phi(\mathbf{I}_f)))],\label{eq:g_loss}
\end{align}
where $\mathbf{I}_r$ and $\mathbf{I}_f$ are real data and generated data, respectively. The third term in \cref{eq:d_loss} is the gradient penalty, and $\lambda$ denotes the weight for this term.

\noindent{\textbf{Symmetry Loss}}. Planar underlying shapes are easily generated when single-view images are synthesized for training. To avoid the trivial solution, we ask the network to synthesize the second image under another view, which we choose the symmetrical camera view $\xi'$ regarding the $yz$-plane,
\begin{align}
    \mathbf{I}_f' = \mathrm{G}(\mathbf{z}, \xi').
\end{align}
The novel view image will be used to calculate the adversarial loss as in \cref{eq:d_loss} and \cref{eq:g_loss}, and get $\mathcal{L}_{\mathrm{D}'}$ and $\mathcal{L}_{\mathrm{G}'}$.

\noindent\textbf{Pose Loss.}
As stated in \cref{sec:method:discriminator}, we also attach an auxiliary pose branch $\Phi(\cdot)$ in the discriminator to perform pose estimation on the given images.
We use \cref{eq:pose-loss} for pose branch training.

\noindent\textbf{Full Objectives.}
In summary, the pose-free generator and the pose-aware discriminator are jointly optimized with
\begin{align}
    \mathcal{L} = & \mathcal{L}_\mathrm{G} + \mathcal{L}_{\mathrm{G}'}
                 + \mathcal{L}_\mathrm{D}+ \mathcal{L}_{\mathrm{D}'} + \gamma\mathcal{L}_{pose},
\end{align}
where $\gamma$ is the weight of pose loss.

\subsection{Implementation Details}

We build our \method on the architecture of EG3D~\cite{eg3d}.
For the pose-aware generator, we do not use pose-conditioned generation. Instead, we instantiate the pose branch with two linear layers and a leaky ReLU activation in between. This branch takes in latent codes from $w$ space and outputs camera poses.
The camera pose consists of an azimuth angle and an elevation angle, which indicates the camera position for rendering.
The triplane resolution is $256\times 256$, and the rendering in the neural radiance field is conducted on $64\times64$ resolution. Both the feature map and the image are rendered.
A super-resolution module then transforms the feature map into a high-resolution image.
The pose-aware discriminator is inherited from the dual discriminator in EG3D~\cite{eg3d}.
We add the pose branch on the features before the last two fully-connected layers that output the realness score. The pose branch is composed of two fully-connected layers with a leaky ReLU as the activation function in the middle.
The learning rate of the generator is $2.5e$-$3$ while that of its pose branch is set to $2.5e$-$5$.
The discriminator's learning rate is $2e$-$3$.
More details are available in the \supp.

%% file: sections/4.exp.tex
\section{Experiments}\label{sec:exp}

\subsection{Experimental Settings}
\noindent\textbf{Datasets.}
We evaluate \method on three datasets, including FFHQ~\cite{stylegan}, Cats~\cite{cats}, and Shapenet Cars~\cite{shapenet}.
FFHQ is a real-world dataset that contains unique 70K high-quality images of human faces.
We follow \cite{eg3d} to align and crop the data.
Cats includes 10K real-world cat images of various resolutions. The data is preprocessed following \cite{gram}.
Shapenet Cars~\cite{shapenet} is a synthetic dataset that contains different car models. We use the dataset rendered from \cite{eg3d} that has around 530K images. Unlike the face-forward datasets, the camera poses of it span the entire 360$^{\circ}$ azimuth and 180$^{\circ}$ elevation distributions.
In the experiments, we use the resolution of $256\times256$ for FFHQ and Cats, and $128\times128$ for Shapenet Cars.

\begin{figure*}[t]
\begin{center}
\includegraphics[width=1\linewidth]{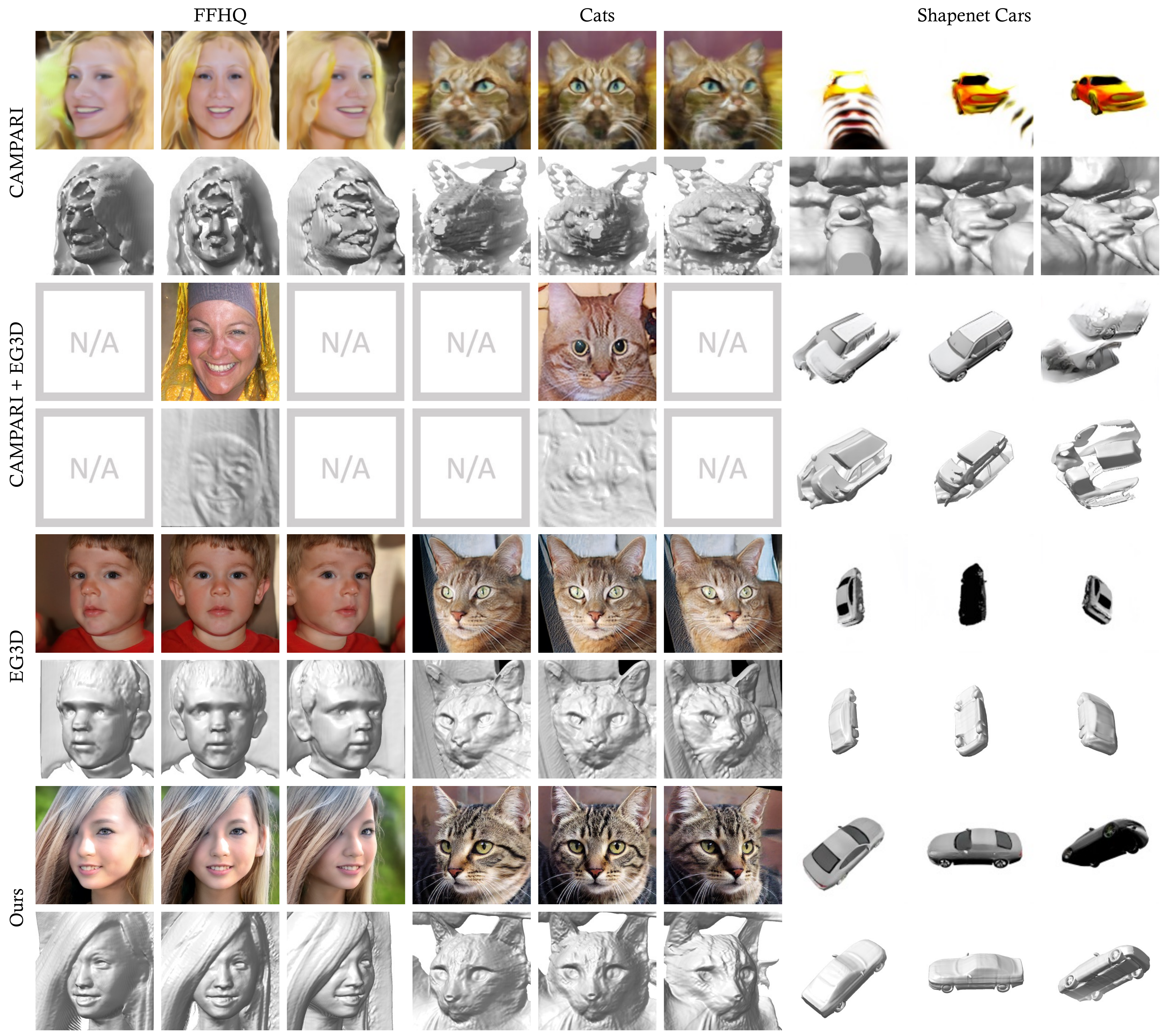}
\end{center}
\vspace{-20pt}
\caption{\textbf{Qualitative comparison between our approach and baselines.} CAMPARI~\cite{campari} struggles to generate reasonable results. ``CAMPARI + EG3D'' suffers from mode collapse on the FFHQ and Cats datasets. EG3D~\cite{eg3d} achieves impressive rendering and reconstruction results when camera poses of training images are given. In contrast, our approach can generate high-quality rendering and geometry without any pose priors.}
\vspace{-15pt}
\label{fig:qualitative}
\end{figure*}

\noindent\textbf{Baselines.}
We compare our approach against two methods: CAMPARI~\cite{campari}, the state-of-the-art pose learning method in 3D-aware image synthesis, and EG3D~\cite{eg3d}, the state-of-the-art in 3D-aware image synthesis. We also build a baseline method which is a combination of them, where we incorporate the pose learning method in CAMPARI into the framework of EG3D.
More details regarding the baselines can be found in \supp.

\begin{figure*}[t]
\begin{center}
\includegraphics[width=0.87\linewidth]{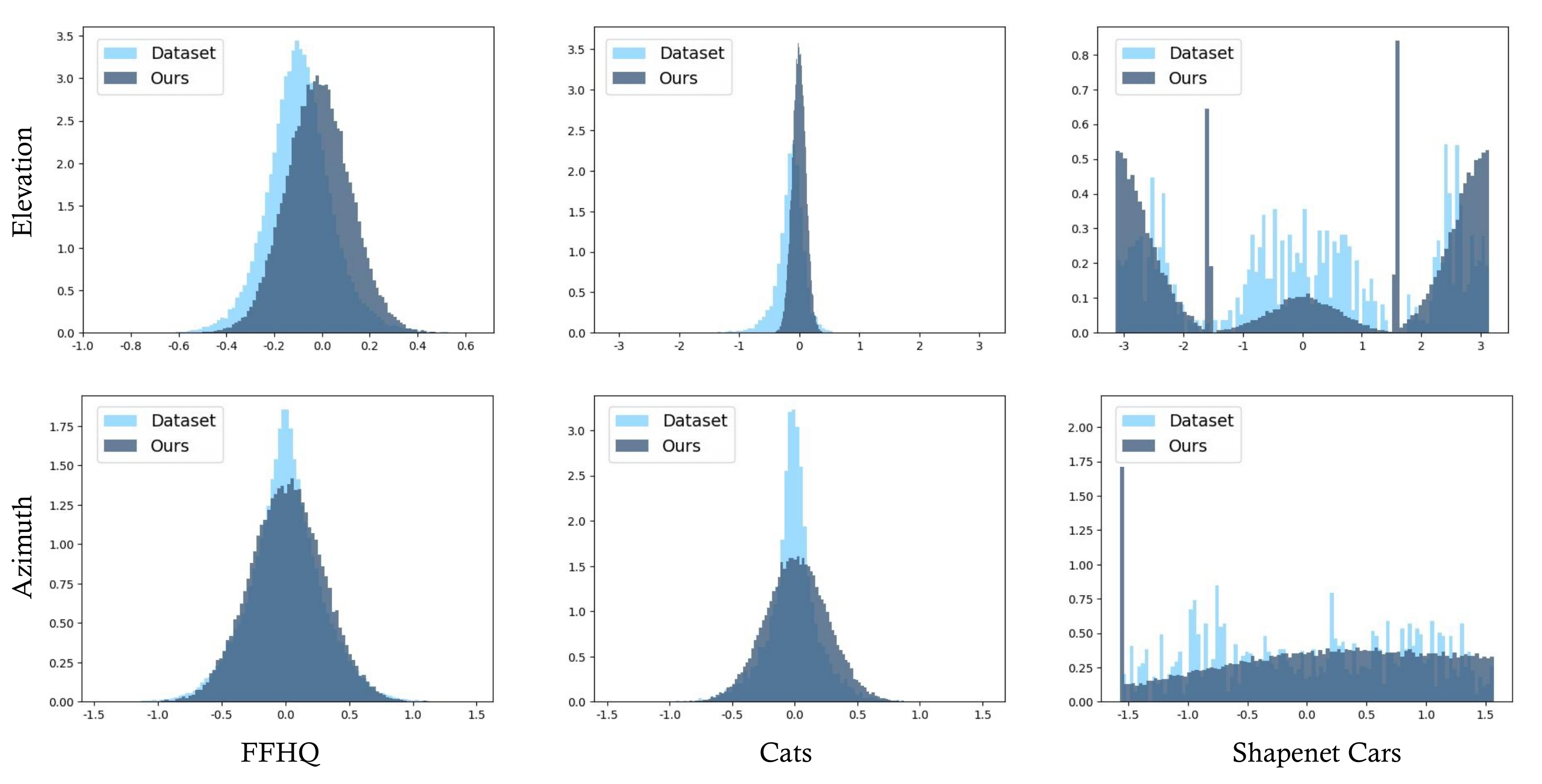}
\end{center}
\vspace{-20pt}
\caption{\textbf{Pose distribution learned by the pose-free generator.} The proposed approach well captures the pose distribution of the dataset. Note that, there is a slight shift of elevation distribution on the FFHQ dataset. The reason is that the model can automatically learn a canonical space for best performance during training.}
\vspace{-10pt}
\label{fig:generatorpose}
\end{figure*}

\noindent\textbf{Metrics.}
We use five metrics to evaluate the performance, including Fr\'{e}chet Inception Distance (FID)~\cite{fid}, Depth Error~\cite{eg3d}, Pose Error~\cite{volumegan}, Reprojection Error (RE)~\cite{volumegan}, and Jensen–Shannon Divergence (JS).
FID is measured between 50K generated images and all real images.
Depth Error is used to assess the quality of geometry. We follow \cite{eg3d} and calculate the mean squared error (MSE) against pseudo-ground-truth depth estimated by \cite{deng2019accurate} on 10K synthesized samples.
We evaluate the pose accuracy on 10K generated samples as well. Given the synthesized images, we leverage the head pose estimator~\cite{whenet} and report the L1 distance between the estimated poses and the poses inferred from the generator. Note that we subtract the respective means of estimated and inferred pose distribution to compensate the canonical shift problem (see \cref{fig:generatorpose}).
Inter-view consistency is measured by the re-projection error following \cite{volumegan}. We render 5 views by sampling azimuth uniformly in the range of [-23$^\circ$, 23$^\circ$] and warp two consecutive views to each other to report the MSE. The images are normalized in the range of [-1, 1] for evaluation. Since the azimuth range learned by CAMPARI is much smaller, we use the learned range for sampling in the measurement of CAMPARI.
To measure the quality of the learned pose distribution, we report Jensen–Shannon Divergence on 50K synthesized samples and all real images by averaging divergence values over azimuth and elevation.

\setlength{\tabcolsep}{7pt}
\begin{table*}
\center
\caption{
    \textbf{Quantitative comparison} on FFHQ~\cite{stylegan}, Cats~\cite{cats}, and Shapenet Cars~\cite{shapenet}. Our approach significantly outperforms CAMPARI~\cite{campari} and ``CAMPARI + EG3D'' in terms of depth error, pose error, reprojection error, and Jensen-Shannon divergence score. Note that, EG3D~\cite{eg3d} uses pose annotations, while our model is trained without any pose priors. * means we report the FID of EG3D on Shapenet Cars using its released model, which is inconsistent with the one reported in the paper.
}
\vspace{-10pt}
\begin{tabular}{lcccccccccc}
\toprule
\multirow{2}{*}{Model} & \multicolumn{5}{c}{FFHQ~\cite{stylegan}} & \multicolumn{3}{c}{Cats~\cite{cats}} & \multicolumn{2}{c}{Shapenet Cars~\cite{shapenet}} \\\cmidrule(lr){2-6} \cmidrule(lr){7-9} \cmidrule(lr){10-11}

& FID$_\text{50k}\downarrow$ & Depth$\downarrow$ &  Pose$\downarrow$ & RE$\downarrow$ & JS$\downarrow$ & FID$_\text{50k}\downarrow$ & RE$\downarrow$& JS$\downarrow$ & FID$_\text{50k}\downarrow$ & JS$\downarrow$ \\
\midrule
CAMPARI~\cite{campari} & 58.59 & 1.78 & 0.15 & 0.109 & 0.61 & 37.40 & 0.050 & 0.60 & 68.91 & 0.72 \\ 
CAMPARI+EG3D & \bf{3.25} & 1.13 & 0.18 & $-$ & 0.73 & \bf{4.83} & $-$ & 0.74 & 4.66 & 0.83\\
EG3D~\cite{eg3d} & 4.80 & \bf{0.29} & \bf{0.07} & 0.039 & $-$ & 5.56 & \bf{0.042} & $-$ & 9.68* & $-$\\
\midrule
Ours & 4.99 & \bf{0.29} & 0.10 & \bf{0.037} & \bf{0.20} & 5.46 & 0.046 & \bf{0.24} & \bf{3.78}  &  \bf{0.51}  \\
\bottomrule
\end{tabular}
\vspace{-15pt}
\label{tab:quantitative}
\end{table*}

\subsection{Main Results}\label{sec:mainresults}
\noindent\textbf{Qualitative Comparison.}
\cref{fig:qualitative} shows the qualitative comparison against the baselines.
CAMPARI learns faulty and very sharp shapes that span the entire space along the ray direction (the same phenomenon as observed in \cref{fig:teaser}d). Therefore, even if the object is rotated with an extremely small angle (e.g., 2.5$^\circ$), the caused visual effect is equivalent to that of the normal rotation of a decent shape with a much larger angle (e.g., 23$^\circ$).
We visualize CAMPARI in its own valid range of the horizontal angle ( around [-2.5$^\circ$, 2.5$^\circ$] for FFHQ, [-5$^\circ$, 5$^\circ$] for Cats, and [-30$^\circ$, 30$^\circ$] for Shapenet Cars).
For others, visualizations on FFHQ and Cats are conducted on the angle range [-23$^\circ$, 23$^\circ$]. Poses for cars are randomly sampled from the entire 360$^{\circ}$ azimuth and 180$^{\circ}$ elevation distributions.
EG3D leverages ground-truth camera poses for training and generates high-quality images and underlying geometry simultaneously. 
When incorporated with the pose learning method in CAMPARI, EG3D suffers from pose collapse and converges to one pose on FFHQ and Cats. Hence, novel views are not available in such cases, and the 3D-aware image synthesis problem is degraded to 2D image synthesis. 
On Shapenet Cars, the generated image and its corresponding shape are visually reasonable under a specific view but become completely unnatural when deviating from that view.
Our method, without any pose priors, can generate high-fidelity images and shapes on par with EG3D.

\noindent\textbf{Quantitative Comparison.}
We report the quantitative evaluation of baselines and our method in \cref{tab:quantitative}.
CAMPARI fails to capture the underlying pose distribution and gets high Jensen–Shannon Divergence scores and pose errors. 
Consequently, the quality of images and shapes is unsatisfying as well, reflected by high FID scores and depth errors.
CAMPARI+EG3D generates planar shapes and collapses to one pose or a small range of poses, which simplifies 3D-aware image synthesis to 2D image synthesis, resulting in lower FID scores but higher depth error, pose error and Jensen–Shannon Divergence score.
EG3D, the method that uses pose annotations, exhibits good performance in terms of image quality and geometry quality.
Our method yields results on par with EG3D and learns much better underlying pose distribution than the baselines.
The quality of the geometry affects the multi-view consistency of the learned neural fields. Thus, EG3D and ours, which synthesize better shapes, can get better scores of reprojection errors.
Note that our pose error is higher than that of EG3D, one of the reasons is that there exists the canonical view shift. As we do not provide any information on what the canonical view will be like, the model tends to learn a canonical view that leads to the best of its performance, which could be different from the standard. More discussions are in \cref{sec:analysis}.

\begin{figure*}[t]
\begin{center}
\includegraphics[width=0.87\linewidth]{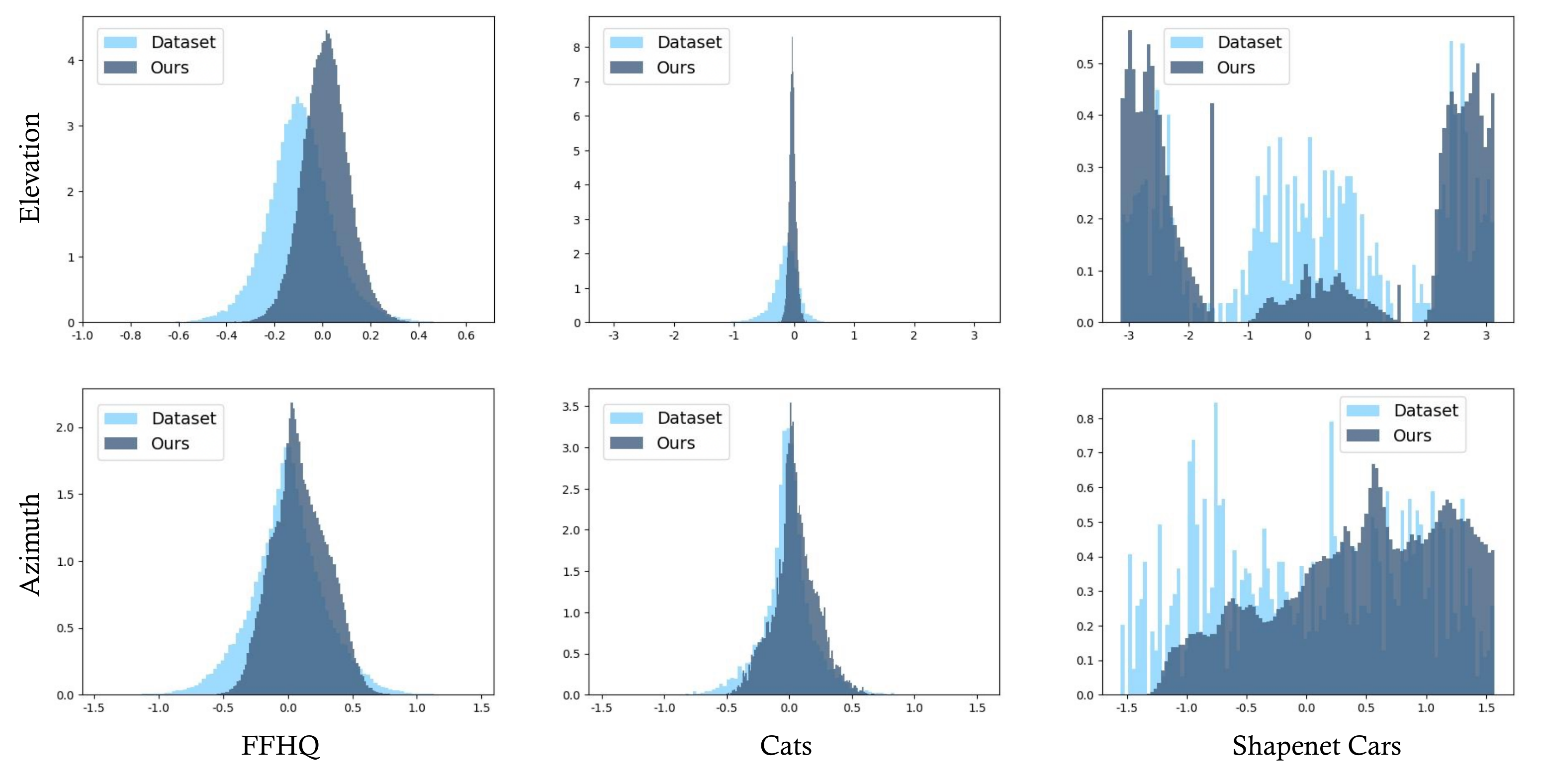}
\end{center}
\vspace{-20pt}
\caption{\textbf{Pose distribution learned by the pose-aware discriminator.} Results demonstrate that our discriminator can learn the pose distribution of the dataset. The issue of canonical space shift also exists from the discriminator perspective, same as \cref{fig:generatorpose}.}
\vspace{-10pt}
\label{fig:discriminatorpose}
\end{figure*}

\setlength{\tabcolsep}{2.5pt}
\begin{table}
\center
\caption{\textbf{Ablation studies} on FFHQ dataset~\cite{stylegan}. We analyze the influence of symmetry loss, pose-aware discriminator, and learning rate. See Sec.~\ref{sec:ablation} for more details.}
\vspace{-10pt}
\begin{tabular}{lccccc}
\toprule
\multirow{2}{*}{Model} & \multicolumn{4}{c}{FFHQ} \\ \cmidrule(lr){2-5}

& FID$_\text{50k}$ $\downarrow$ & Depth$\downarrow$ &  Pose$\downarrow$ & JS$\downarrow$ \\
\midrule
\textit{w/o} symmetry loss & 4.50 & 0.41 & 0.12  & 0.20  \\ 
\textit{w/o} pose-aware D & \bf{3.43} & 1.30 & 0.19  & 0.21 \\ 
\textit{w/o} pose condition in D & 3.51 & 1.48 & 0.19 & 0.26 \\
lr $=2.5e$-$4$ & 122.07 & 0.82 & 0.74 & 0.56 \\ %
lr $=2.5e$-$6$ & 10.20 & 0.70 & 0.16 & 0.38 \\
\midrule
Ours & 4.99 & \bf{0.29} & \bf{0.10} & \bf{0.20}  \\ 
\bottomrule
\end{tabular}
\vspace{-18pt}
\label{tab:ablation}
\end{table}

\subsection{Ablation Study}\label{sec:ablation}
We conduct ablation studies on FFHQ $256\times 256$ to analyze the effectiveness of \method.
Evaluation results are shown in \cref{tab:ablation}.

\noindent\textbf{Symmetry Loss.}
We render an image from the symmetrical camera view and leverage it to pose a constraint on the generated neural radiance field from another view. 
Without this constraint, although the network can synthesize visually pleasing images under the learned pose and capture a rough pose distribution, the underlying geometry has a tendency to flatten, resulting in higher depth error.
The constraint from the symmetrical view can detect the flat shapes because the rendered image under the novel view deviates from the real distribution and shall be corrected by the model.

\begin{figure*}[t]
\begin{center}
\includegraphics[width=1\linewidth]{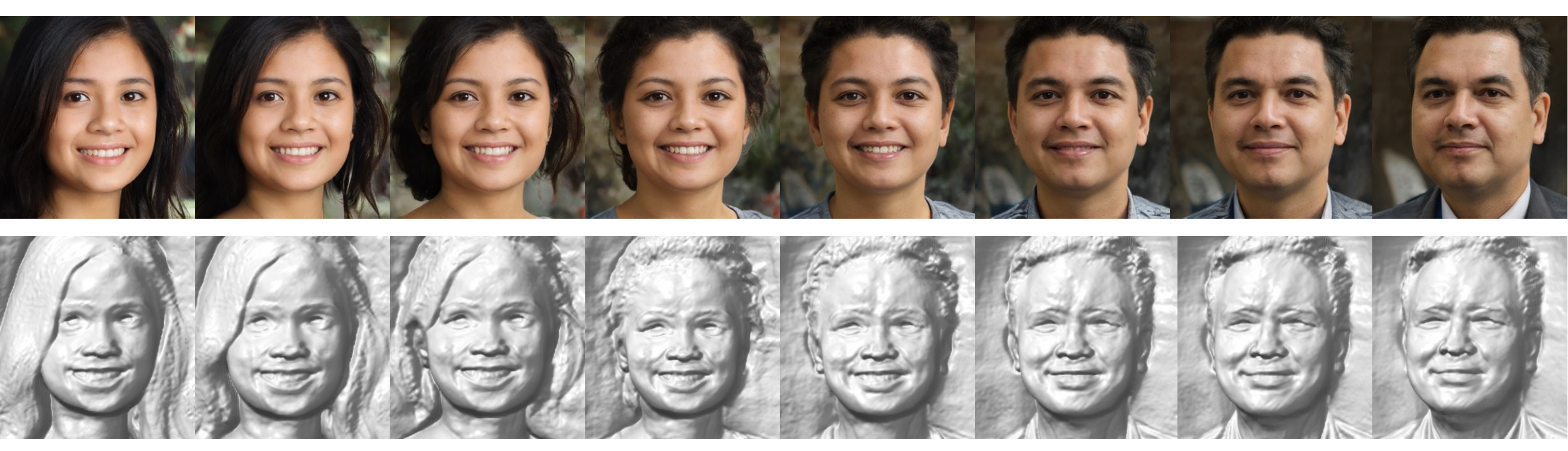}
\end{center}
\vspace{-20pt}
\caption{\textbf{Results of linear interpolation between two latent codes.} The appearance, the geometry, and the pose all change smoothly, indicating the latent space is semantically meaningful.}
\vspace{-15pt}
\label{fig:interpolation}
\end{figure*}

\noindent\textbf{Pose-aware Discriminator.}
To verify the capability of the pose-aware discriminator, we compare it against the simple dual discriminator that removes the conditional label and purely performs real or fake discrimination.
The simple dual discriminator does not provide any additional pose information to the generator other than the common realness score. Thus, the generator learns the pose freely and has difficulty aligning poses of data to share the same canonical view. 
For example, in the human face generation, one fake pose can correspond to images under different real poses. As a result, the pose error increases, and planar shapes are more frequently generated.
We also report the results where only the pose conditioning discrimination is changed to the conventional pose-free discrimination. It drastically harms the geometry quality.
With our pose-aware discriminator, the discriminator predicts the pose from both the fake and real data and uses it as the condition, which forces the generator to synthesize images under the same view as the conditional label, leading to better pose alignment.

\noindent\textbf{Learning Rate of Pose Learner.}
Learning rate of the pose branch in the generator matters a lot.
With a large learning rate (\textit{e.g.,} $2.5e$-$4$ in \cref{tab:ablation}), the generator always tries with new poses of large variation at the beginning and leaves the content learning behind, resulting in unstable training and higher tendency of mode collapse.
As reported in \cref{tab:ablation}, all the evaluation metrics experience a significant drop.
On the contrary, if the learning rate is too small (\textit{e.g.,} $2.5e$-$6$ in \cref{tab:ablation}), the generator cannot explore the pose distribution enough and easily gets stuck in a small range of pose distribution, putting more focus on the content learning. 
Consequently, the model fails to capture and understand the pose distribution, leading to worse shapes.
A proper learning rate should balance the exploration of pose distribution and the learning of the content.

\subsection{Pose Analysis}\label{sec:analysis}
In this section, we analyze the pose distributions that are learned by our pose-free generator and pose-aware discriminator.

\noindent\textbf{Poses in Pose-free Generator.}
To visualize the pose distribution learned in the pose-free generator, we randomly sample 50K latent codes and infer the poses from them. 
The pose distributions from the dataset and learned in the generator are shown in \cref{fig:generatorpose}.
Without any pose priors, our method can capture the pose range and distribution of the dataset in general.
As observed in the figure for elevation distributions on the FFHQ dataset, there exists a distribution shift. 
The reason is that we do not release any information about the canonical view to the model, and therefore, the model can interpret the canonical view freely with different angles which might be different from the one defined in the dataset.
Generally, the model will choose the one that leads to the best performance as its canonical view.

\noindent\textbf{Poses in Pose-aware Discriminator.}
Since our discriminator is equipped with a pose predictor, we hire it to predict the poses for all data in the dataset and draw the distributions accordingly in \cref{fig:discriminatorpose}.
For FFHQ and Cats datasets, the distributions are well approximated, and the canonical view shift problem also exists in the discriminator.
On Shapenet Cars, the distribution of azimuth is biased to one side. We conjecture the reason is that the model has difficulty distinguishing the front and rear of the car as they look very similar. 
As a result, the learned distribution shifts towards a hemisphere instead of the entire sphere.

\subsection{Applications}

\noindent\textbf{Linearity in Latent Space.}
To demonstrate the latent space learned by \method is semantically meaningful, we randomly sample two latent codes and linearly interpolate between them. 
The interpolation results are shown in \cref{fig:interpolation}.
We can see that both the appearance and the underlying geometry are changing smoothly. 
Besides, as we infer the pose from the latent code, poses undergo smooth changes as well.

\begin{figure}[t]
\begin{center}
\includegraphics[width=1\linewidth]{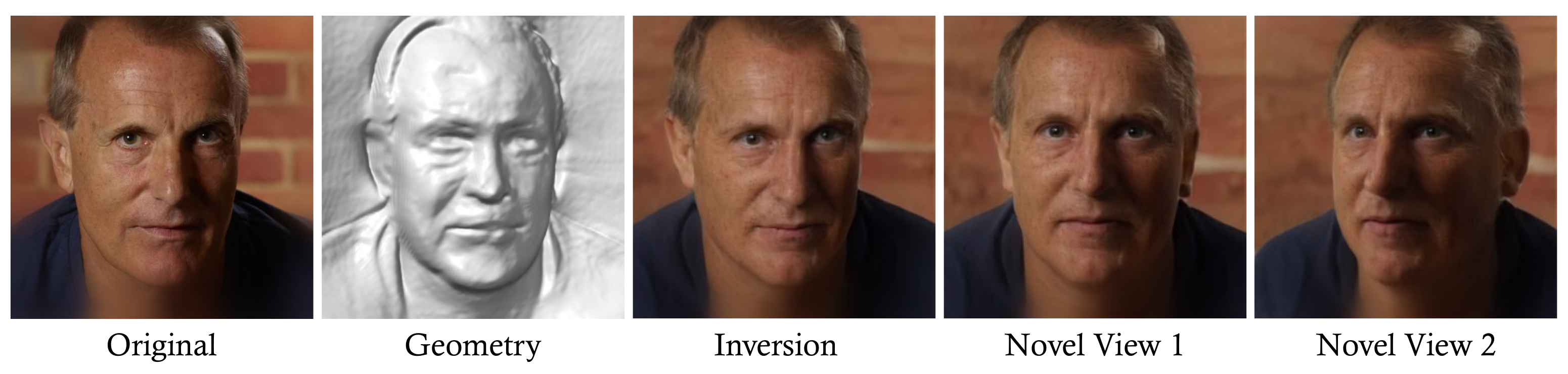}
\end{center}
\vspace{-20pt}
\caption{\textbf{Results of GAN inversion.} Thanks to the design of pose-free generator, we do not require pose estimation prior to performing inversion like previous works. Instead, we can directly invert the image to get the 3D representation and the corresponding pose. The resconstructed geometry is realistic and enables vivid novel view synthesis.}
\vspace{-10pt}
\label{fig:inversion}
\end{figure}

\noindent\textbf{Real Image Inversion.}
One of the potential applications of \method is to reconstruct the geometry from a real image and enable novel view synthesis.
To achieve this goal, PTI~\cite{PTI}, a GAN inversion method, is adopted to perform on our pose-free generator.
In the previous methods~\cite{eg3d} that demonstrate GAN inversion as an application, an off-the-shelf pose predictor is required to obtain the pose of the given image prior to performing GAN inversion.
Ours, however, benefits from the changed formulation in which the camera pose is encoded in the latent space. Therefore, we can get not only the inverted neural radiance field but also the camera pose under which the given image is captured, when performing GAN inversion.
The result is shown in \cref{fig:inversion}.
Our method can successfully invert the image and synthesize vivid images under novel views.

%% file: sections/5.conclusion.tex
\section{Conclusion}\label{sec:conclusion}

This work presents \method, which learns 3D-aware image synthesis without using any pose priors.
A pose-free generator is designed to infer camera poses directly from the latent space so that the requirement for pose priors in the previous works is removed.
Besides, to help the generator better capture the underlying distribution, we make the discriminator pose-aware by asking it to first predict the camera pose and then use it as a condition when performing conditional real/fake discrimination.
With such a design, experimental results demonstrate that our approach is able to synthesize high-quality images and high-fidelity shapes that are on par with state-of-the-art 3D synthesis methods without the need for manual pose tuning or dataset labeling.

\noindent\textbf{Acknowledgement.} This research has been made possible by funding support from the Research Grants Council of Hong Kong under the Research Impact Fund project R6003-21.

%% file: sections/6.ref.tex
{\small
\bibliographystyle{ieee_fullname}
\bibliography{ref}
}

%% file: supp/0.overview.tex
\clearpage
\appendix
\section*{Appendix}

In this appendix, we first provide the training details of our \method in \cref{sec:training}.
In \cref{sec:baselines}, we describe the details of implementations of baselines.
\cref{sec:results} provides more qualitative results.
Moreover, we show the syntheses under steep angles.
\cref{sec:discussion} discusses the limitations as well as the potential future work of \method.
Ethical considerations are also provided.

%% file: supp/1.training.tex
\section{Training and Implementation Details}\label{sec:training}

\noindent\textbf{Training Details.} Most of our training parameters are the same as those in EG3D~\cite{eg3d}.
We reset the loss weight $\lambda$ for gradient penalty to 1.0, 5.0, 0.3 for FFHQ~\cite{stylegan}, Cats~\cite{cats}, and Shapenet Cars~\cite{shapenet}, respectively.
$\gamma$, the weight for pose loss, is set to 2, 10, 2 for FFHQ, Cats, and Shapenet Cars. 
All losses are used for training iteratively.
For FFHQ and Cats, models are trained on the NeRF resolution of $64\times 64$ and the image resolution of $256\times256$.
While for models on Shapenet Cars, the NeRF resolution is $64\times 64$ and the image resolution of $128\times128$, following the setting in \cite{eg3d}.
Models on FFHQ and Shapenet Cars are trained end-to-end on 25000K images for around 6 days on 8 NVIDIA A100 GPUs.
Due to the limited amount of data in Cats dataset, we follow the setting in EG3D~\cite{eg3d} to finetune the pretrained model of FFHQ on Cats dataset for 600K images.

\noindent\textbf{Additional Implementation Details.}
We would like to illustrate more implementation details in addtion to details in Sec.3.5.
PoF3D is built upon EG3D~\cite{eg3d}, including the triplane generator, decoder, volume rendering, super-resolution module and dual discriminator. In the triplane generator, we disable the pose conditioning and add a pose learner. 
The pose learner consists of two linear layers with hidden size 512 and a leaky ReLU in between. It takes in a $w$-space code of size 512 and outputs camera poses of dimension 2, an azimuth angle and an elevation angle.
In the dual discriminator, we add a pose predictor. The pose predictor has the same structure as the pose learner except that the hidden size is 4096 and the input is feature maps of resolution 4 in the discriminator.

%% file: supp/2.baselines.tex
\section{Baselines}\label{sec:baselines}

\noindent\textbf{CAMPARI}~\cite{campari} is a 3D-aware image synthesis method that models camera distribution during training.
We use the \href{https://github.com/autonomousvision/campari}{official implementation} for all experiments. 
For FFHQ, Cats and Shapenet Cars dataset, we keep the settings identical to the provided configurations for CelebA, Cats and Carla, but we allow the learning of azimuth and elevation angle only. Following \cite{campari}, the prior distribution is set to Gaussian distribution $\mathcal{N}(0,13.5^\circ)$ for azimuth and elevation on FFHQ and Cats, and a uniform distribution over the entire azimuth and elevation for Shapenet Cars.  Other camera parameters are fixed to the one learned in the original settings.
Besides, we follow the original setting that the camera distribution will be fixed for later stages of training on FFHQ.

\begin{figure*}[t]
\begin{center}
\includegraphics[width=1\linewidth]{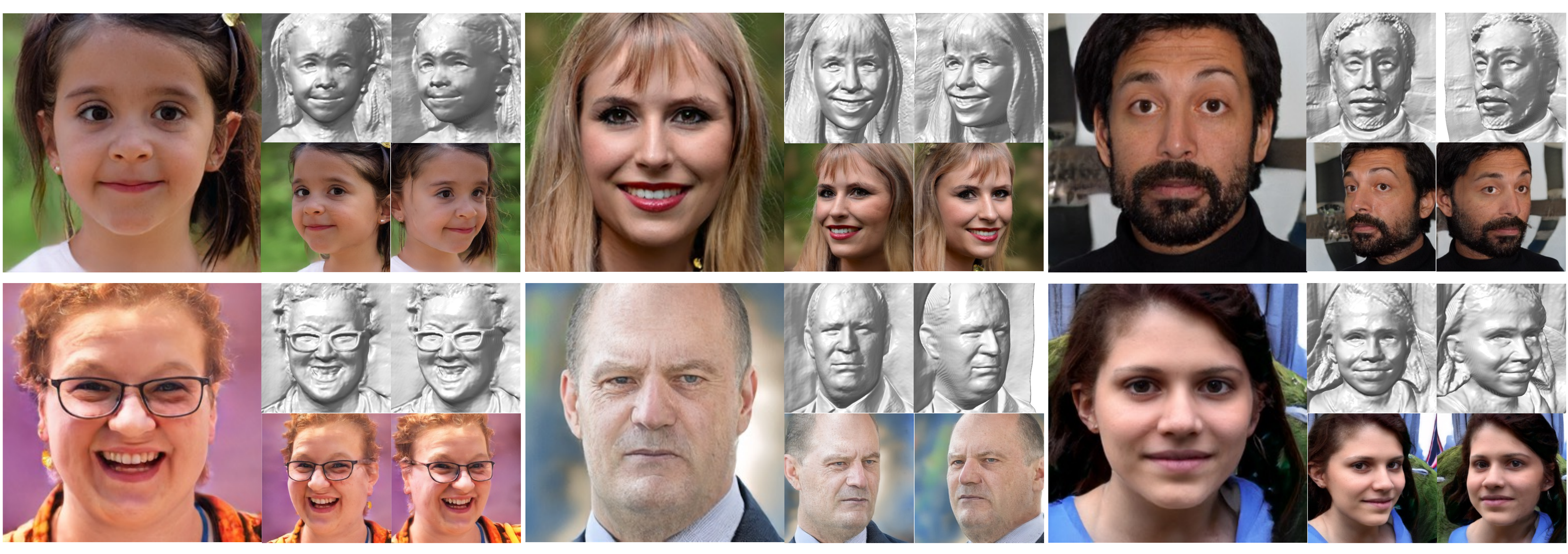}
\end{center}
\vspace{-20pt}
\caption{\textbf{Untruncated samples on FFHQ~\cite{stylegan}.} For each generated identity, we show the underlying geometry under two views and appearance under three views.}
\label{fig:supp_ffhq}
\end{figure*}

\begin{figure*}[t]
\begin{center}
\includegraphics[width=1\linewidth]{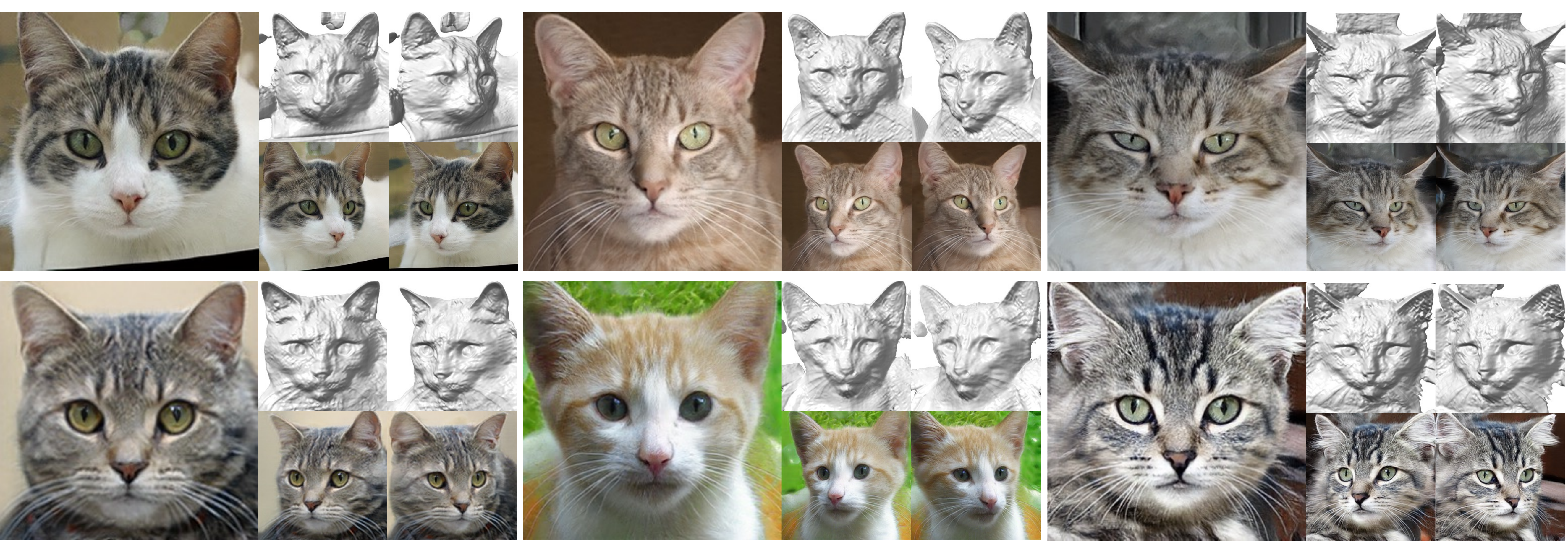}
\end{center}
\vspace{-20pt}
\caption{\textbf{Synthesized samples on Cats~\cite{cats} with truncation 0.7.} For each generated cat, we show the underlying geometry under two views and appearance under three views.}
\label{fig:supp_cats}
\end{figure*}

\begin{figure*}[t]
\begin{center}
\includegraphics[width=1\linewidth]{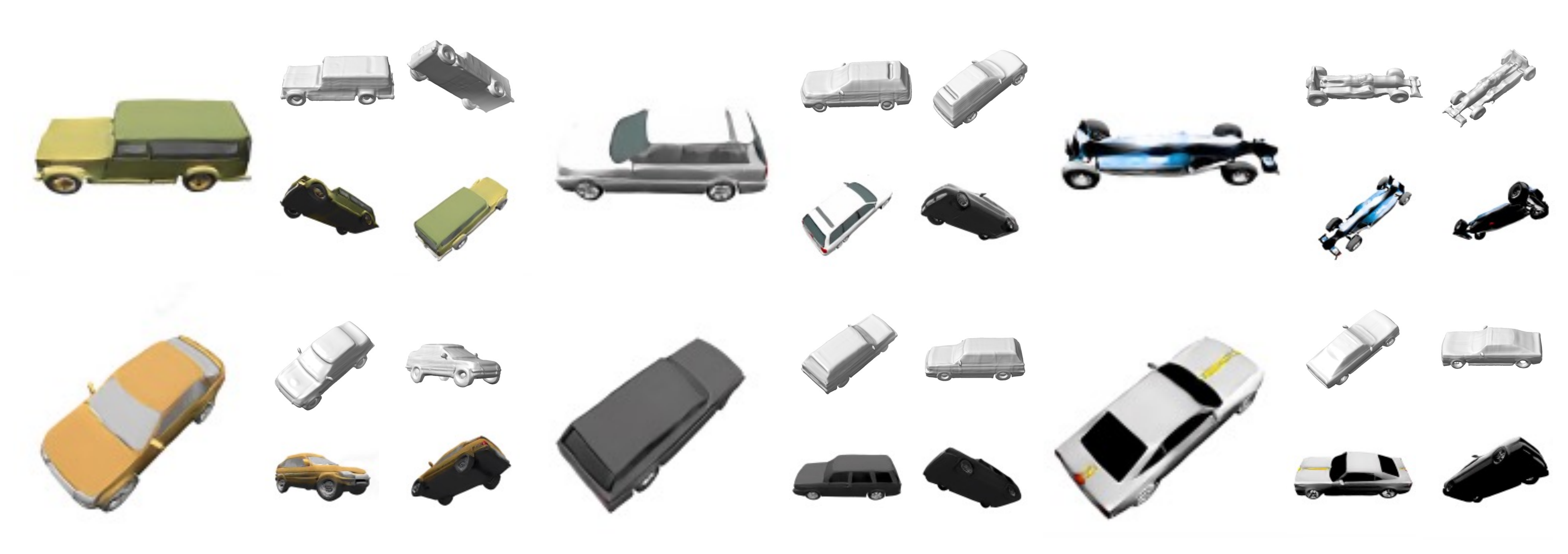}
\end{center}
\caption{\textbf{Synthesized samples on Shapenet Cars~\cite{shapenet} with truncation 0.7}. For each generated car, we show the underlying geometry under two views and appearance under three views.}
\label{fig:supp_cars}
\end{figure*}

\noindent\textbf{EG3D}~\cite{eg3d} is also one of the state-of-the-art methods in 3D-aware image synthesis, which leverages ground-truth camera poses for training.
We use the \href{https://github.com/NVlabs/eg3d}{official implementation} for all experiments.
For FFHQ dataset, since the checkpoint for $256\times256$ has not been released yet, we use the provided configuration to train on the NeRF resolution of $64\times 64$ and image resolution of $256\times 256$.
For Cats dataset, we make use of the pose annotations processed by \cite{gram}. Other settings are identical to the original one for cat dataset, and the model is trained on the NeRF resolution of $64\times 64$ and image resolution of $256\times 256$ as well.
Moreover, we follow \cite{eg3d} to finetune the model with the checkpoint of FFHQ on Cats dataset rather than train the model from scratch.
We adopt the checkpoint of Shapenet Cars provided by the authors for evaluation.

\noindent\textbf{CAMPARI+EG3D} is a combination of CAMPARI~\cite{campari} and EG3D~\cite{eg3d}, where the pose distribution learning network in CAMPARI is merged into the framework of EG3D. 
Concretely, in EG3D, we do not sample poses from the collection of real poses for generation, but sample a pose from a prior distribution and transform it into a proper one with a network. The transformed pose is then used for rendering. For real data, we still leverage the ground-truth poses for conditioning.
The training strategy and the initialization of priors for pose learning in CAMPARI+EG3D follows those in CAMPARI.
Other parameters such as camera intrinsic matrix are identical to those used in EG3D.

%% file: supp/3.results.tex
\section{Additional Results and Analysis}\label{sec:results}

\subsection{Qualitative Results}

We provide more qualitative results in \cref{fig:supp_ffhq,fig:supp_cats,fig:supp_cars}.
A \href{https://www.youtube.com/watch?v=nvlyAElC8eE}{demo video}, is also available to show the qualitative comparison with baselines.
Our results are on par with those generated from EG3D~\cite{eg3d} and much better than those from CAMPARI~\cite{campari}.

\subsection{Syntheses under Steep Angles}

\begin{figure*}[t]
\begin{center}
\includegraphics[width=1\linewidth]{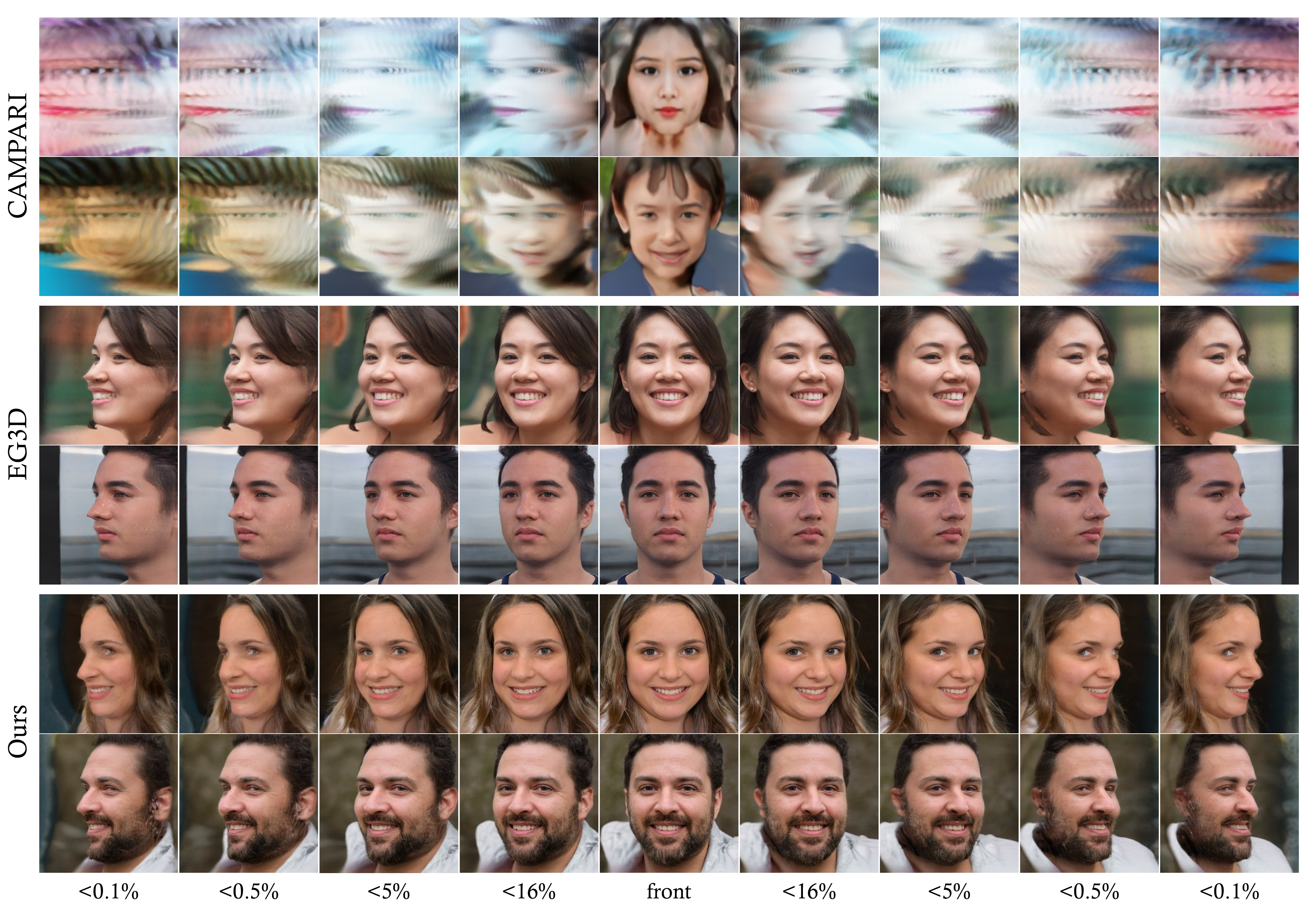}
\end{center}
\caption{\textbf{Syntheses under steep angles.} $<X\%$ denotes less than $X$ percent of training cases are trained under that pose.}
\label{fig:extreme}
\end{figure*}

We synthesize images under steep camera poses on FFHQ dataset~\cite{stylegan} in \cref{fig:extreme}.
Since CAMPARI fails to learn a proper pose distribution and generates sharp and bumpy shapes as discussed in Sec. 4.2, it finds it hard to synthesize reasonable images under larger rotation.
EG3D leverages ground-truth poses for training and is good at generating images under extreme views. However, it tends to generate extremely sharp noses.
Ours, however, can synthesize natural noses and high-quality images under steep angles without using any pose prior.

\subsection{Training Behavior}
\begin{figure*}[!ht]
\begin{center}
\includegraphics[width=1\linewidth]{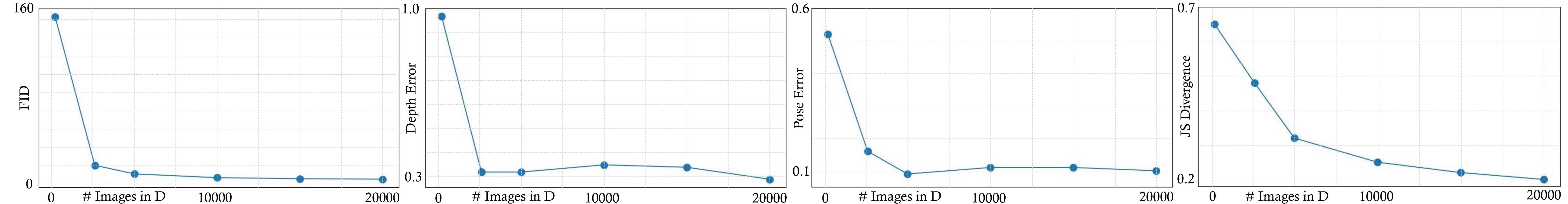}
\end{center}
\caption{\textbf{Behavior as training progresses.} Zoom in for details.}
\label{fig:trainingprogress}
\end{figure*}

We show the trends of FID, depth error, pose error and Jensen-Shannon divergence in \cref{fig:trainingprogress} as training progresses. Generally, the network learns fast at first and slows down later. The learning of the data distribution is slower than the other three aspects.

\subsection{3D Reconstruction using COLMAP}
We render 128 views from a random code using the same camera trajectory as ~\cite{eg3d}, to reconstruct a point cloud using COLMAP. As shown in \cref{fig:reconstruction}, the \textit{dense} point cloud indicates the good multi-view consistency achieved by PoF3D.

\begin{figure*}[!ht]
\begin{center}
\includegraphics[width=1\linewidth]{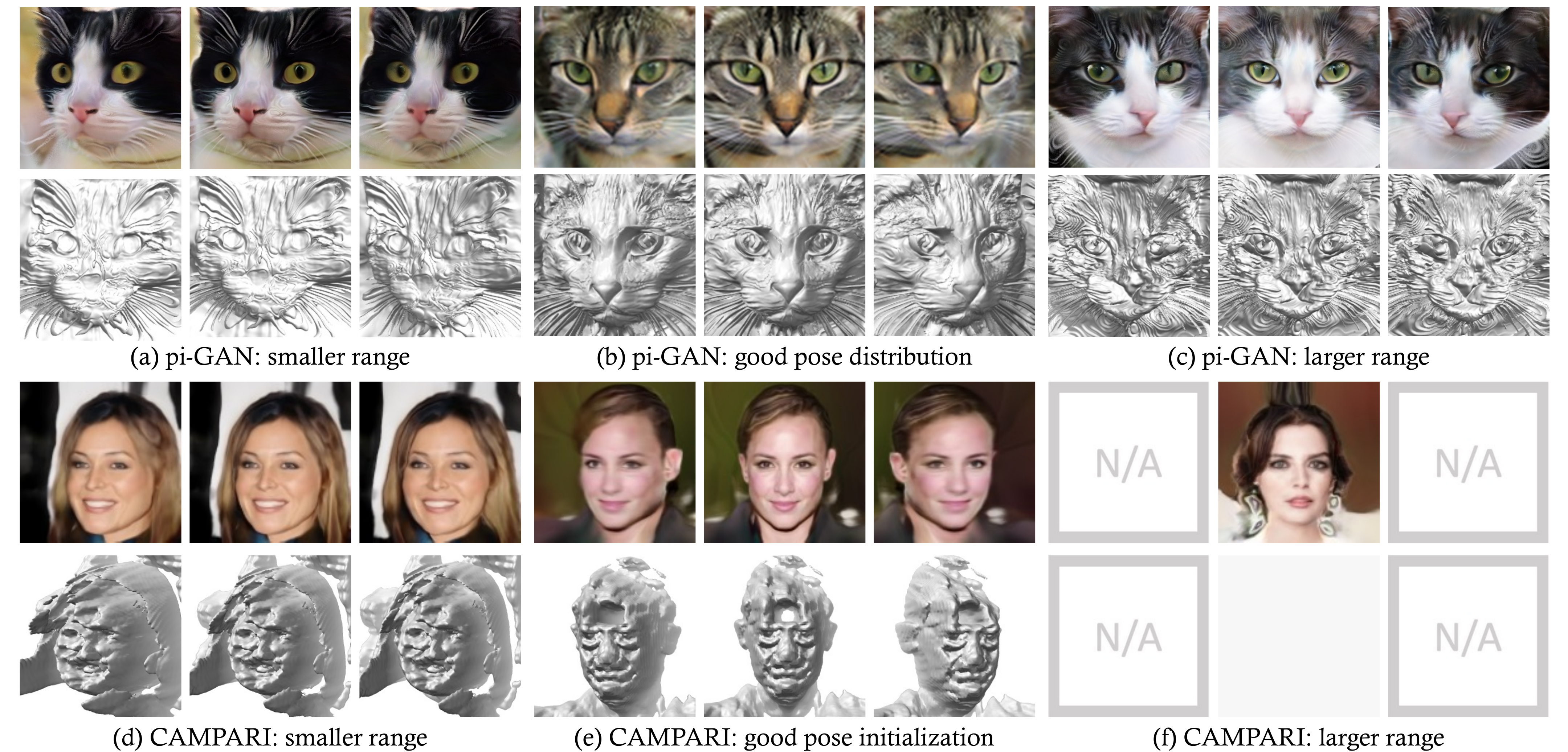}
\end{center}
\caption{\textbf{Results of different pose priors.} FID scores of (a) to (f) are 12, 17, 13, 36, 28, and 26.}
\label{fig:teaserre}
\end{figure*}

\begin{figure}[!ht]
\begin{center}
\includegraphics[width=1\linewidth]{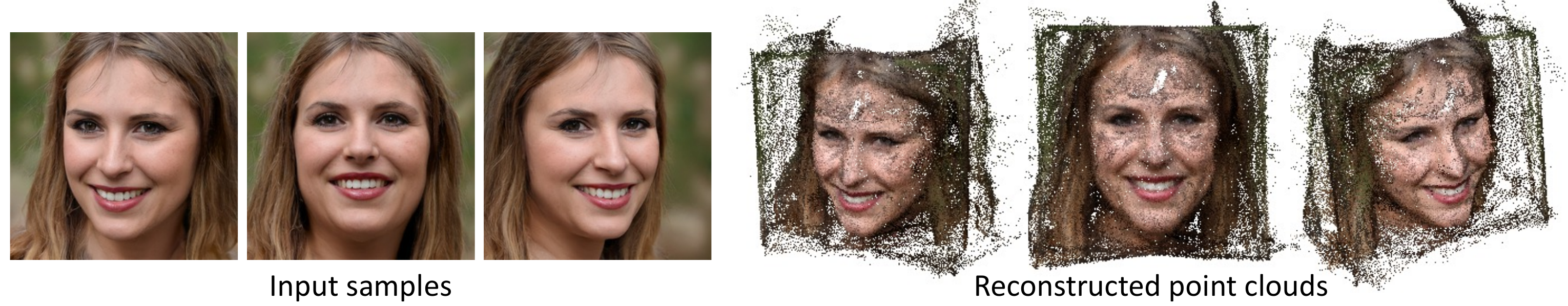}
\end{center}
\caption{\textbf{3D reconstruction using COLMAP}.}
\label{fig:reconstruction}
\end{figure}

\subsection{Distribution Difference between G and D}
\begin{figure}[!ht]
\begin{center}
\includegraphics[width=1\linewidth]{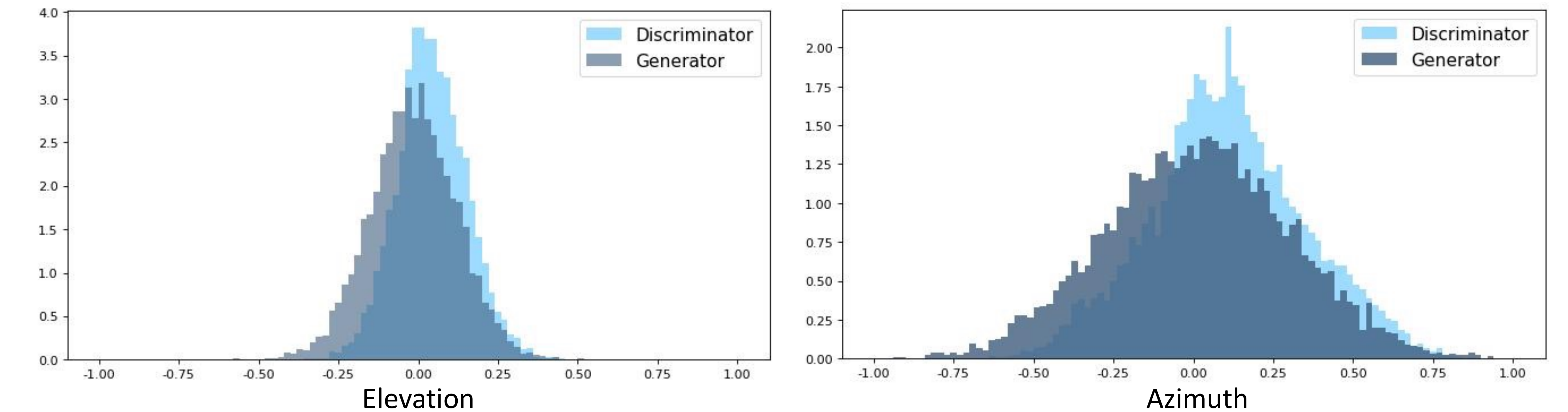}
\end{center}
\caption{\textbf{Pose distributions in G and D} trained on FFHQ. }
\label{fig:difference}
\end{figure}

\cref{fig:difference} visualizes the distribution discrepancy of G and D on FFHQ, where the pose error is 0.09.
The reason for the distribution discrepancy is that in GAN training, it is hard to optimize to the optimal point. A sub-optimal solution brings the difference on pose distributions in G and D, as well as the non-zero FID. How to make G and D equivalent is a long-standing problem.

\subsection{Full analysis of Fig. 1}
We provide the full analysis of baselines in Fig. 1 with both smaller range of pose distribution and larger range of pose distribution, showing how sensitive existing works are to the pre-estimated pose prior.
As shown in \cref{fig:teaserre}, for $\pi$-GAN (top), with [-0.5, 0.5] (middle) as the optimal prior, using [-0.3, 0.3] (left) and [-0.7, 0.7] (right) result in \textit{(i)} planar and noisy shape as well as \textit{(ii)} the loss of canonical space.
Similarly, for CAMPARI (bottom), with 0.24 (middle) as the optimal pose std, using 0.12 (left) and 0.36 (right) harm the performance drastically.

%% file: supp/4.discussions.tex
\section{Discussion}\label{sec:discussion}
\subsection{Limitations and Future Work}

Though \method generates high-quality images and decent underlying shapes without pose priors, there are still some artifacts on the geometry.
For example, the eye balls have concave underlying shapes, leading to incorrect movement during rotation.
We believe extra geometry supervision shall be added on them to fix the problem.
Sometimes bumpy regions can be observed. We think with larger batch size, the pose distribution can be learnt more accurately and thus leads to more decent shapes.
Texture sticking effect is also noticed during rotation, which might be mitigated by replacing the StyleGAN2 backbone with StyleGAN3~\cite{stylegan3}.

Despite the well-captured pose distribution, \method sometimes confuses the front with the rear of the car.
The reason is that the front and the rear of cars look similar to each other in Shapenet Cars~\cite{shapenet}, a synthetic dataset.
A more powerful pose predictor should be introduced into the discriminator to improve the ability of judging the front and the rear of cars, which we leave for future work.

We do not model the foreground and the background separately, and thus the background is close to the foreground objects from time to time.
Techniques, such as NeRF++~\cite{nerf++}, can be integrated into our framework to model the foreground and background independently, which is also a potential future direction to be explored.

\subsection{Ethical Considerations}
\method can benefit vision and graphics applications, such as gaming and content creation.
However, it also poses a threat because generative models can be misused for DeepFake-related applications, \textit{e.g.}, human face editing and talking head generation.
We hope that DeepFake detection algorithms can be developed to avoid such misuse.
In addition, verification cues, such as forensics, offer another solution to mitigate the problem.

%% file: main.bbl
\begin{thebibliography}{10}\itemsep=-1pt

\bibitem{3DGP}
Anonymous.
\newblock 3d generation on imagenet.
\newblock In {\em Submitted to The Eleventh International Conference on
  Learning Representations}, 2023.
\newblock under review.

\bibitem{mipnerf}
Jonathan~T Barron, Ben Mildenhall, Matthew Tancik, Peter Hedman, Ricardo
  Martin-Brualla, and Pratul~P Srinivasan.
\newblock Mip-nerf: A multiscale representation for anti-aliasing neural
  radiance fields.
\newblock In {\em Int. Conf. Comput. Vis.}, pages 5855--5864, 2021.

\bibitem{eg3d}
Eric~R Chan, Connor~Z Lin, Matthew~A Chan, Koki Nagano, Boxiao Pan, Shalini
  De~Mello, Orazio Gallo, Leonidas~J Guibas, Jonathan Tremblay, Sameh Khamis,
  et~al.
\newblock Efficient geometry-aware 3d generative adversarial networks.
\newblock In {\em IEEE Conf. Comput. Vis. Pattern Recog.}, 2022.

\bibitem{pigan}
Eric~R Chan, Marco Monteiro, Petr Kellnhofer, Jiajun Wu, and Gordon Wetzstein.
\newblock pi-gan: Periodic implicit generative adversarial networks for
  3d-aware image synthesis.
\newblock In {\em IEEE Conf. Comput. Vis. Pattern Recog.}, 2021.

\bibitem{shapenet}
Angel~X Chang, Thomas Funkhouser, Leonidas Guibas, Pat Hanrahan, Qixing Huang,
  Zimo Li, Silvio Savarese, Manolis Savva, Shuran Song, Hao Su, et~al.
\newblock Shapenet: An information-rich 3d model repository.
\newblock {\em arXiv preprint arXiv:1512.03012}, 2015.

\bibitem{gram}
Yu Deng, Jiaolong Yang, Jianfeng Xiang, and Xin Tong.
\newblock Gram: Generative radiance manifolds for 3d-aware image generation.
\newblock In {\em IEEE Conf. Comput. Vis. Pattern Recog.}, 2022.

\bibitem{deng2019accurate}
Yu Deng, Jiaolong Yang, Sicheng Xu, Dong Chen, Yunde Jia, and Xin Tong.
\newblock Accurate 3d face reconstruction with weakly-supervised learning: From
  single image to image set.
\newblock In {\em IEEE Conf. Comput. Vis. Pattern Recog. Worksh.}, 2019.

\bibitem{gan}
Ian Goodfellow, Jean Pouget-Abadie, Mehdi Mirza, Bing Xu, David Warde-Farley,
  Sherjil Ozair, Aaron Courville, and Yoshua Bengio.
\newblock Generative adversarial nets.
\newblock In {\em Adv. Neural Inform. Process. Syst.}, 2014.

\bibitem{stylenerf}
Jiatao Gu, Lingjie Liu, Peng Wang, and Christian Theobalt.
\newblock Stylenerf: A style-based 3d-aware generator for high-resolution image
  synthesis.
\newblock {\em arXiv preprint arXiv:2110.08985}, 2021.

\bibitem{fid}
Martin Heusel, Hubert Ramsauer, Thomas Unterthiner, Bernhard Nessler, and Sepp
  Hochreiter.
\newblock Gans trained by a two time-scale update rule converge to a local nash
  equilibrium.
\newblock In {\em Adv. Neural Inform. Process. Syst.}, 2017.

\bibitem{stylegan3}
Tero Karras, Miika Aittala, Samuli Laine, Erik H\"ark\"onen, Janne Hellsten,
  Jaakko Lehtinen, and Timo Aila.
\newblock Alias-free generative adversarial networks.
\newblock In {\em Adv. Neural Inform. Process. Syst.}, 2021.

\bibitem{stylegan}
Tero Karras, Samuli Laine, and Timo Aila.
\newblock A style-based generator architecture for generative adversarial
  networks.
\newblock In {\em IEEE Conf. Comput. Vis. Pattern Recog.}, 2019.

\bibitem{stylegan2}
Tero Karras, Samuli Laine, Miika Aittala, Janne Hellsten, Jaakko Lehtinen, and
  Timo Aila.
\newblock Analyzing and improving the image quality of {StyleGAN}.
\newblock In {\em IEEE Conf. Comput. Vis. Pattern Recog.}, 2020.

\bibitem{kuang2022neroic}
Zhengfei Kuang, Kyle Olszewski, Menglei Chai, Zeng Huang, Panos Achlioptas, and
  Sergey Tulyakov.
\newblock Neroic: Neural rendering of objects from online image collections.
\newblock {\em arXiv preprint arXiv:2201.02533}, 2022.

\bibitem{celeba}
Ziwei Liu, Ping Luo, Xiaogang Wang, and Xiaoou Tang.
\newblock Deep learning face attributes in the wild.
\newblock In {\em Int. Conf. Comput. Vis.}, 2015.

\bibitem{occupancy}
Lars Mescheder, Michael Oechsle, Michael Niemeyer, Sebastian Nowozin, and
  Andreas Geiger.
\newblock Occupancy networks: Learning 3d reconstruction in function space.
\newblock In {\em IEEE Conf. Comput. Vis. Pattern Recog.}, 2019.

\bibitem{nerf}
Ben Mildenhall, Pratul~P Srinivasan, Matthew Tancik, Jonathan~T Barron, Ravi
  Ramamoorthi, and Ren Ng.
\newblock Nerf: Representing scenes as neural radiance fields for view
  synthesis.
\newblock In {\em Eur. Conf. Comput. Vis.}, 2020.

\bibitem{hologan}
Thu Nguyen-Phuoc, Chuan Li, Lucas Theis, Christian Richardt, and Yong-Liang
  Yang.
\newblock Hologan: Unsupervised learning of 3d representations from natural
  images.
\newblock In {\em Int. Conf. Comput. Vis.}, 2019.

\bibitem{nguyen2020blockgan}
Thu~H Nguyen-Phuoc, Christian Richardt, Long Mai, Yongliang Yang, and Niloy
  Mitra.
\newblock Blockgan: Learning 3d object-aware scene representations from
  unlabelled images.
\newblock {\em Adv. Neural Inform. Process. Syst.}, 2020.

\bibitem{campari}
Michael Niemeyer and Andreas Geiger.
\newblock Campari: Camera-aware decomposed generative neural radiance fields.
\newblock In {\em International Conference on 3D Vision (3DV)}, 2021.

\bibitem{giraffe}
Michael Niemeyer and Andreas Geiger.
\newblock Giraffe: Representing scenes as compositional generative neural
  feature fields.
\newblock In {\em IEEE Conf. Comput. Vis. Pattern Recog.}, 2021.

\bibitem{stylesdf}
Roy Or-El, Xuan Luo, Mengyi Shan, Eli Shechtman, Jeong~Joon Park, and Ira
  Kemelmacher-Shlizerman.
\newblock Stylesdf: High-resolution 3d-consistent image and geometry
  generation.
\newblock In {\em IEEE Conf. Comput. Vis. Pattern Recog.}, 2022.

\bibitem{shadegan}
Xingang Pan, Xudong Xu, Chen~Change Loy, Christian Theobalt, and Bo Dai.
\newblock A shading-guided generative implicit model for shape-accurate
  3d-aware image synthesis.
\newblock In {\em Adv. Neural Inform. Process. Syst.}, 2021.

\bibitem{deepsdf}
Jeong~Joon Park, Peter Florence, Julian Straub, Richard Newcombe, and Steven
  Lovegrove.
\newblock Deepsdf: Learning continuous signed distance functions for shape
  representation.
\newblock In {\em IEEE Conf. Comput. Vis. Pattern Recog.}, 2019.

\bibitem{neuralbody}
Sida Peng, Yuanqing Zhang, Yinghao Xu, Qianqian Wang, Qing Shuai, Hujun Bao,
  and Xiaowei Zhou.
\newblock Neural body: Implicit neural representations with structured latent
  codes for novel view synthesis of dynamic humans.
\newblock In {\em IEEE Conf. Comput. Vis. Pattern Recog.}, pages 9054--9063,
  2021.

\bibitem{PTI}
Daniel Roich, Ron Mokady, Amit~H Bermano, and Daniel Cohen-Or.
\newblock Pivotal tuning for latent-based editing of real images.
\newblock {\em ACM Trans. Graph.}, 2021.

\bibitem{SFM}
Johannes~L Schonberger and Jan-Michael Frahm.
\newblock Structure-from-motion revisited.
\newblock In {\em IEEE Conf. Comput. Vis. Pattern Recog.}, 2016.

\bibitem{graf}
Katja Schwarz, Yiyi Liao, Michael Niemeyer, and Andreas Geiger.
\newblock Graf: Generative radiance fields for 3d-aware image synthesis.
\newblock In {\em Adv. Neural Inform. Process. Syst.}, 2020.

\bibitem{voxgraf}
Katja Schwarz, Axel Sauer, Michael Niemeyer, Yiyi Liao, and Andreas Geiger.
\newblock Voxgraf: Fast 3d-aware image synthesis with sparse voxel grids.
\newblock {\em Adv. Neural Inform. Process. Syst.}, 2022.

\bibitem{MVS}
Steven~M Seitz, Brian Curless, James Diebel, Daniel Scharstein, and Richard
  Szeliski.
\newblock A comparison and evaluation of multi-view stereo reconstruction
  algorithms.
\newblock In {\em IEEE Conf. Comput. Vis. Pattern Recog.}, 2006.

\bibitem{shi2022deep}
Zifan Shi, Sida Peng, Yinghao Xu, Yiyi Liao, and Yujun Shen.
\newblock Deep generative models on 3d representations: A survey.
\newblock {\em arXiv preprint arXiv:2210.15663}, 2022.

\bibitem{depthgan}
Zifan Shi, Yujun Shen, Jiapeng Zhu, Dit{-}Yan Yeung, and Qifeng Chen.
\newblock 3d-aware indoor scene synthesis with depth priors.
\newblock In {\em Eur. Conf. Comput. Vis.}, 2022.

\bibitem{shi2022improving}
Zifan Shi, Yinghao Xu, Yujun Shen, Deli Zhao, Qifeng Chen, and Dit-Yan Yeung.
\newblock Improving 3d-aware image synthesis with a geometry-aware
  discriminator.
\newblock In {\em Adv. Neural Inform. Process. Syst.}, 2022.

\bibitem{epigraf}
Ivan Skorokhodov, Sergey Tulyakov, Yiqun Wang, and Peter Wonka.
\newblock Epigraf: Rethinking training of 3d gans.
\newblock In {\em Adv. Neural Inform. Process. Syst.}, 2022.

\bibitem{wang2021nerfmm}
Zirui Wang, Shangzhe Wu, Weidi Xie, Min Chen, and Victor~Adrian Prisacariu.
\newblock Ne{RF}$--$: Neural radiance fields without known camera parameters.
\newblock {\em arXiv preprint arXiv:2102.07064}, 2021.

\bibitem{xia2022survey}
Weihao Xia and Jing-Hao Xue.
\newblock A survey on 3d-aware image synthesis.
\newblock {\em arXiv preprint arXiv:2210.14267}, 2022.

\bibitem{gof}
Xudong Xu, Xingang Pan, Dahua Lin, and Bo Dai.
\newblock Generative occupancy fields for 3d surface-aware image synthesis.
\newblock In {\em Adv. Neural Inform. Process. Syst.}, 2021.

\bibitem{xu2022discoscene}
Yinghao Xu, Menglei Chai, Zifan Shi, Sida Peng, Ivan Skorokhodov, Aliaksandr
  Siarohin, Ceyuan Yang, Yujun Shen, Hsin-Ying Lee, Bolei Zhou, et~al.
\newblock Discoscene: Spatially disentangled generative radiance fields for
  controllable 3d-aware scene synthesis.
\newblock {\em arXiv preprint arXiv:2212.11984}, 2022.

\bibitem{volumegan}
Yinghao Xu, Sida Peng, Ceyuan Yang, Yujun Shen, and Bolei Zhou.
\newblock 3d-aware image synthesis via learning structural and textural
  representations.
\newblock In {\em IEEE Conf. Comput. Vis. Pattern Recog.}, 2022.

\bibitem{ghfeat}
Yinghao Xu, Yujun Shen, Jiapeng Zhu, Ceyuan Yang, and Bolei Zhou.
\newblock Generative hierarchical features from synthesizing images.
\newblock In {\em IEEE Conf. Comput. Vis. Pattern Recog.}, 2021.

\bibitem{nerf++}
Kai Zhang, Gernot Riegler, Noah Snavely, and Vladlen Koltun.
\newblock Nerf++: Analyzing and improving neural radiance fields.
\newblock {\em arXiv preprint arXiv:2010.07492}, 2020.

\bibitem{cats}
Weiwei Zhang, Jian Sun, and Xiaoou Tang.
\newblock Cat head detection-how to effectively exploit shape and texture
  features.
\newblock In {\em Eur. Conf. Comput. Vis.}, 2008.

\bibitem{stylempi}
Xiaoming Zhao, Fangchang Ma, David G{\"u}era, Zhile Ren, Alexander~G Schwing,
  and Alex Colburn.
\newblock Generative multiplane images: Making a 2d gan 3d-aware.
\newblock In {\em Eur. Conf. Comput. Vis.}, 2022.

\bibitem{whenet}
Yijun Zhou and James Gregson.
\newblock Whenet: Real-time fine-grained estimation for wide range head pose.
\newblock In {\em Brit. Mach. Vis. Conf.}, 2020.

\bibitem{von}
Jun-Yan Zhu, Zhoutong Zhang, Chengkai Zhang, Jiajun Wu, Antonio Torralba,
  Joshua~B. Tenenbaum, and William~T. Freeman.
\newblock Visual object networks: Image generation with disentangled 3{D}
  representations.
\newblock In {\em Adv. Neural Inform. Process. Syst.}, 2018.

\end{thebibliography}
